# Streamlining Admission with LOR Insights: AI-Based Leadership Assessment in Online Master's Program

Meryem Yilmaz Soylu[1], Adrian Gallard[1], Jeonghyun Lee[1], Gayane Grigoryan[1], Rushil Desai[1], Stephen Harmon[1]

[1] Georgia Institute of Technology, United States

Corresponding author: Meryem Yilmaz Soylu, meryem@gatech.edu

## Abstract

Letters of recommendation (LORs) provide valuable insights into candidates' capabilities and experiences beyond standardized test scores. However, reviewing these text-heavy materials is time-consuming and labor-intensive. To address this challenge and support the admission committee in providing feedback for students' professional growth, our study introduces LORI: LOR Insights, a novel AI-based detection tool for assessing leadership skills in LORs submitted by online master's program applicants. By employing natural language processing and leveraging large language models by using RoBERTa and LLAMA, we seek to identify leadership attributes such as teamwork, communication, and innovation. Our latest RoBERTa model achieves a weighted F1 score of 91.6%, a precision of 92.4%, and a recall of 91.6%, showing a strong level of consistency in our test data. With the growing importance of leadership skills in the STEM sector, integrating LORI—a tool designed with cutting-edge AI models—into the graduate admissions process is crucial for accurately assessing applicants' leadership capabilities. This approach not only streamlines the admissions process but also automates and ensures a more comprehensive evaluation of candidates' capabilities.

**Keywords:** leadership, 21st-century skills, durable skills, natural language processing, machine learning, large language models, graduate education, holistic admission

## Introduction

Since the outbreak of the COVID-19 pandemic, it has become clear that various challenges to personal and national economic stability, coupled with rapid advancements in technology and infrastructure, are significantly changing our work and lifestyle dynamics. As a result, these changes are influencing our educational priorities. The increasing need for top-notch education is going beyond conventional school environments and geographical borders, leading to the emergence of online learning platforms that cater to all educational levels and are accessible to learners across the globe. Notably, numerous institutions have recently introduced online graduate degree programs spanning diverse fields. However, while these programs address the growing demand-supply gap, the mere acquisition of subject matter expertise falls short of adequately equipping individuals to navigate and excel in our rapidly evolving societal landscape. As ongoing transformations continue, it's crucial to develop adaptable skills that can help individuals thrive. These skills are generally referred to as 21st-century skills (21CS), which were acknowledged by researchers (e.g., (Bentur et al., 2019; Lavi et al., 2021; Marbach-Ad et al., 2015)), educational institutions (e.g.,(ABET, 2019; Knowledge, 2012; P21, 2019)), and economic organizations (e.g., (Ananiadou & Claro, 2009; Forum, 2016)).

Among 21CS, leadership is a highly valued skill in both professional and academic settings. It is critical for higher education institutions to identify and nurture students exhibiting robust leadership qualities. This is particularly crucial for prospective graduate students, as demonstrating a degree of leadership aptitude is essential in showcasing their potential for future advancement. However, there are almost no standardized methods available to evaluate leadership skills during the graduate student admission process. Typically, an applicant's suitability for a program is evaluated through standardized tests (e.g., GRE) and written application documents such as essays, statements of purpose, or letters of recommendation (LORs). Among these, LORs provide valuable insights from external perspectives regarding applicants' experiences and leadership abilities. Yet, manually scrutinizing these letters to assess such competencies demands significant time and resources. To address this challenge, our goal is to develop an AI-driven tool capable of analyzing LORs submitted for an online master's program (OMP) application, with the objective of identifying indicators of leadership.

**Related Work**

**LORs in the admission process**

Holistic admissions or "whole-file" review is the consideration of the "broad range of candidate qualities, including non-cognitive or personal attributes when reviewing applications for admissions" ((Kent & McCarthy, 2016) p. 1). In holistic admissions, LORs play a significant role, as they offer unique insights into an applicant's personal and professional characteristics and qualities that extend beyond traditional academic metrics like GPA and test scores. This approach helps graduate programs foster diversity by considering a broader range of candidate qualities, aligning with the principles of the Council of Graduate Schools. LORs provide narratives that offer depth to an application, reflecting personal attributes such as leadership, professionalism, and adaptability (Okahana et al., 2018) and are frequently a factor in final admissions decisions (Posselt, 2018).

However, despite their importance, LORs are subject to criticism due to their unstandardized nature (Dalal et al., 2022; Houser & Lemmons, 2018; Kuncel et al., 2014), the variation in the context of the writer (Clinedinst & Koranteng, 2017; McCarthy et al., 2010), and bias from the writer, the reader, or both (Akos & Kretchmar, 2016; Dalal et al., 2022; Grimm et al., 2020; Posselt, 2018; Sagaria, 2002), which can perpetuate inequality.

A recent study of over 31,000 LORs identified content differences based on gender, race, and intersections of both, although these factors beyond GPA and test scores were not predictive of admission outcomes (Dalal et al., 2022). Additionally, Kim et al. (2024) applied advanced natural language processing to examine over 600,000 counselor recommendation letters, finding notable disparities in length and content tied to race, socioeconomic status, and school type, emphasizing the importance of context-sensitive evaluations in the admissions process.

These findings highlight the complexities of selective admissions. Despite inherent biases, LORs remain valuable in the admissions process as they provide crucial insights into applicants' intellectual engagement, creativity, and potential, helping admissions committees differentiate between candidates with similar academic credentials (Butt, 2024). This encourages the development of tools that allow for deeper analysis of LORs to better support admission officers.

**Leadership skills in graduate school and beyond**

Today, most admissions officers report that their institutions use holistic review in their admissions process (Bastedo et al., 2018; Haviland et al., 2023). This approach allows graduate programs to assess various applicant qualities, including academic preparedness, demonstrated interest in a specific field, research experience, and 21CS—alternatively referred to as soft, non-cognitive, durable or lifetime skills, such as leadership and perseverance (Gooch et al., 2024; Michel et al., 2019; Nye & Ryan, 2023; Paris & Birnbaum, 2024; Paris et al., 2020).

Among these skills, leadership development is recognized as a critical objective across all disciplines, especially in STEM fields. Studies show that the most effective leaders not only master technical expertise but also excel in professional skills like communication and collaboration (Akhtar, 2020; Denecke et al., 2017). Globally, business leaders and executives often prioritize leadership and talent development programs, recognizing that individuals with strong leadership abilities are essential for ensuring smooth project execution and the timely completion of tasks (Denecke et al., 2017; Lawrence et al., 2018). For graduate students in the sciences, technical proficiency is a given, while those who possess leadership training are increasingly sought after by employers (Brookes et al., 2017; National Academies of Sciences & Medicine, 2018).

Given the significance of leadership, possessing these skills has become highly advantageous for applicants seeking acceptance into graduate-level programs. Leadership capabilities demonstrate a candidate's ability to collaborate effectively, take initiative, communicate clearly, and solve complex problems, all of which highlight their potential for success in the rigorous academic and professional environments of graduate education (Sandlin, 2019). Moreover, these attributes suggest a candidate's readiness to assume leadership roles within both academic and professional communities, qualities that are highly valued for future career success (Chhinzer & Russo, 2017).

Additionally, research suggests that alignment between applicants' goals and program objectives, along with their demonstrated competencies in 21CS, significantly influences admissions decisions (Walpole et al., 2002). Among these skills, leadership has emerged as a key predictor of not only enrollment but also retention and overall success in graduate programs (Gomez, 2013; Kyoung Ro et al., 2017). As a result, higher education institutions actively seek evidence of these qualities in application materials, including LORs (Hout, 2005; Kuncel et al., 2014; Sandlin, 2019; Sternberg, 2012).

**Leveraging NLP to review LORs**

Examining LORs like text-heavy application materials is a time-consuming and labor-intensive task. However, recent advancements in technology have led to the development of various artificial intelligence (AI) tools capable of analyzing different attributes of applicants efficiently. One notable application is Natural Language Processing (NLP), a specialized application of machine learning (ML) tailored for interpreting natural language data. NLP techniques use a combination of statistical, ML, and deep learning approaches to understand, interpret, and categorize text based on its content, context, and structure (Holdsworth, 2024).

The strength of NLP lies in its ability to transform unstructured human language into structured data that can be analyzed, interpreted, and applied across various contexts. NLP techniques allow for the efficient processing of vast amounts of text data, automating tasks that would otherwise require significant manual effort (Gruetzemacher, 2022). Advanced models, such as Bidirectional Encoder Representations from Transformers (BERT) and Generative Pre-trained Transformers (GPT), are capable of interpreting ambiguous language, understanding idiomatic expressions, and capturing the nuanced meanings of words within their context. This makes NLP especially powerful for tasks like sentiment analysis, machine translation, and text classification (Devlin et al., 2019).

Language is often ambiguous, meaning that the same word or phrase can have different meanings depending on the context. NLP systems are particularly skilled at resolving this ambiguity by identifying the correct meaning based on the surrounding text. This ability allows NLP to effectively interpret homonyms (words with multiple meanings), metaphors, and other complex language structures. These skills are especially valuable for tasks like answering questions and translating text between languages (Haber & Poesio, 2024; Patwardhan et al., 2023). In addition, NLP models can be tailored to specific fields, such as law, medicine, or technical areas. By fine-tuning these models for a particular domain, they become more capable of understanding the unique vocabulary, structure, and nuances found in specialized texts, leading to more accurate and relevant analysis (Bagheri et al., 2023).

NLP has also been explored in the context of education to automate and enhance the analysis of text-heavy educational data to derive insights into improving teaching and learning outcomes. For instance, Authors (2022) contributed to the understanding of cognitive presence in online learning environments by building an ML model that classifies students' discussion forum posts into phases of cognitive presence. By applying a BERT model, their study achieved 92.5% accuracy in predicting cognitive presence. Similarly, Dornauer et al. (2024) developed a German-language cognitive presence classifier for online discussions using linguistic analysis tools, such as Linguistic Inquiry and Word Count (LIWC), and additional learning traces, such as file attachments and course glossary terms. In a recent study (Parker et al., 2024) using a dataset of 2,500 survey comments from biomedical science courses, the authors showed that GPT-4 can achieve human-level performance across various tasks such as classification, extraction, thematic analysis, and sentiment analysis by leveraging effective prompting.

To date, NLP has been utilized to evaluate students' performance in their application materials, notably LORs, for various post-graduate programs, including admission to graduate school (Heilman et al., 2015; Waters & Miikkulainen, 2014), adaptive behavioral compliance (Jeon & Lee, 2023), and predicting neurosurgical residency outcomes (Ortiz et al., 2022). These studies highlighted the important role of LORs in providing crucial insights into applicants' characteristics and backgrounds, which significantly influence admission decisions and subsequent performance in graduate programs. Considering the growing importance of leadership skills in the STEM workforce, integrating NLP methods into the admission process for graduate education programs becomes imperative to assess applicants' leadership competencies accurately. This approach not only makes the admissions process more efficient but also allows for a deeper assessment of candidates' capabilities.

## Methodology

### Leadership Annotation Schema

Reports from The Chronicle of Higher Education and the World Economic Forum emphasize essential 21CS such as leadership, critical thinking, communication, and teamwork (Carlson, 2017; Di Battista et al., 2023). These skills are increasingly in demand, with organizations urged to prioritize their development (Hackett, 2015; Hora, 2019). Leadership development alone accounts for nearly US$50 billion in global investments annually (Deloitte Consulting et al., 2014; Kirchner & Akdere, 2014). Employees who excel in communication, teamwork, and intercultural competence contribute to organizational productivity and retention, and their participation in cross-functional teams further strengthens leadership capabilities (Akdere et al., 2019). Particularly in today's rapidly evolving STEM industries, effective leadership is critical to driving innovation and growth (Akdere et al., 2019; Karimi & Pina, 2021; Lnenicka et al., 2020; Watt, 2003).

Our comprehensive review of leadership training practices in graduate education revealed a wide range of skills incorporated into these programs (Dowsett & Lacey, 2023; Lenhart et al., 2022). Despite the variety of skills covered, the most consistently emphasized were effective communication, teamwork, and innovation (Dowsett & Lacey, 2023; Lenhart et al., 2022; Strubbe et al., 2022) (Figure 1).

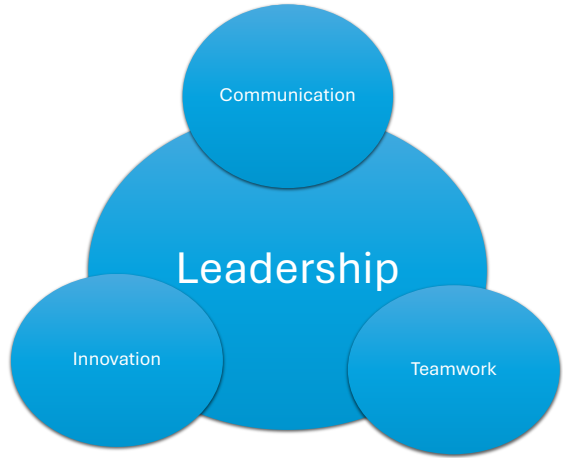

**Fig. 1** Conceptual framework of the leadership in this study

Effective communication, as detailed in various studies (Dowsett & Lacey, 2023; Lenhart et al., 2022), relies on strong listening and comprehension skills, whether in speaking or writing. A key component of this is active listening, which involves paraphrasing the speaker's words, encouraging further elaboration, providing feedback, and ensuring the message is accurately understood. Empathy is also crucial, as it requires receptiveness to others' values and emotions, as well as openly sharing one's own thoughts. When human annotators analyzed the sample of LORs, they looked for language that indicated active listening, the ability to adapt communication to diverse audiences, and strategies for overcoming common communication barriers.

In addition to communication, teamwork is also key to success, as no one can succeed in isolation. Interdepartmental and inter-organizational relationships rely heavily on collaboration (Graesser et al., 2020). Successful collaboration requires openness to diverse perspectives, teamwork in developing plans, and coordinated efforts in execution (García et al., 2016; Koh et al., 2018). As such, human annotators looked for LOR language that highlighted team-building, collaborative work, and the use of tools and platforms to facilitate teamwork.

Finally, innovation lies at the heart of STEM disciplines (Lenhart et al., 2022). It involves questioning the status quo, observing details, and connecting seemingly unrelated concepts. Innovation also requires collaboration with diverse individuals to gain fresh perspectives and experiment with new ideas (Akdere et al., 2019). Accordingly, human annotators sought language that reflected the ability to spot opportunities for innovation, generate and test new ideas through rapid prototyping and user feedback, manage risks, navigate uncertainties, and embrace failure as an essential part of the innovation process.

**Data Collection and Processing**

**Data source**

Data used in this study was gathered from the application packages submitted to the OMP offered by a technology-focused public research university in the U.S. The program, which is designed to improve learners' knowledge of big data analytics techniques through a one-to-two-year program, received more than 10,000 applications as of Spring 2023. The OMP requires the submission of at least three LORs during the application.

Three distinct datasets were prepared for this project. To begin, we required an ample dataset comprising sentences from multiple LORs that were accurately annotated for leadership skills. To obtain this dataset, we employed a Python script to extract individual sentences from the random sample of LORs. Initially, an expert manually annotated sets of LORs from 25 randomly chosen students. Upon analysis of these annotations, we discovered that the dataset was imbalanced with a much larger number of non-leadership sentences than leadership sentences. After generating an initial model that utilized BERT, we applied BERT to a portion of the unlabeled dataset to help our team locate more sentences containing leadership to generate a balanced dataset. By examining the predicted leadership labels and having our expert review and determine which annotations were correct, we were able to include additional leadership sentences to our dataset.

This process resulted in 1,048 sentences from LORs corresponding to 120 unique applicant IDs. These applicants were randomly selected from the entire pool of individuals who applied, regardless of whether they were admitted to the program. The sample of LORs included recommendations written by the applicants' former or current managers, instructors, and colleagues, with the letters varying in format—some were lengthy and detailed, while others were shorter and more informal. This initial set of annotated sentences comprised the first dataset.

These annotated sentences are used to train the weak-supervision models, which utilize datasets where only a portion of the data is manually labeled. This approach leverages a combination of labeled and unlabeled data, making it more cost-effective and efficient compared to fully supervised learning, where all data must be manually labeled (Ratner et al., 2020; Zhou, 2018; Zhu et al., 2023). This data was divided into 943 lines of training data and 105 lines of validation data. In the final model run, the 1048-lined dataset was divided into a second dataset of two equal parts comprising a validation set of 524 and a test set of 524 lines of data. The final datasets were created to provide a larger pool of data for validation and testing for the final model.

The second dataset refers to the processed weakly-labeled dataset produced after running the weak-supervision pipeline. Any overlapping student IDs from the first dataset were removed from the unlabeled dataset. Using weak-supervision techniques, we created over 250,000 lines of data, forming the foundation for training a subsequent weakly-supervised model. Initially, the raw data contained 15,293 unique student IDs and 39,465 distinct LORs. Ultimately, the data for training the final model was machine-annotated, while the previously human-annotated dataset served as a benchmark for validating and testing the weakly-supervised model.

A separate group of LORs from a set of students was pulled from the unlabeled dataset (unique to the sentences of the previous dataset) to form a third dataset to check the inter-rater operability between humans and the ML model. Two experts analyzed these sentences using a library of phrases and keywords associated with leadership skills, including teamwork, communication, and innovation (Author, 2024). The sentences were then labeled with "1" if the leadership skill was present and "0" if not. Based on the predicted label for leadership, the human coders' inter-rater reliability, measured using Cohen's Kappa, was 0.65, indicating a substantial level of agreement among the raters (Kolesnyk & Khairova, 2022; Landis & Koch, 2016; Sun, 2011).

**Preprocessing**

Preprocessing steps were conducted at both the weak-supervision pipeline development and model training stages to ensure data quality and enhance performance. These steps included handling outliers, generating numeric features with the Spacy library, and using regex for text pattern matching and word separation functions. The Spacy library was used for NLP tasks such as tokenization and feature extraction, while regex helped identify and manage specific text patterns during data cleaning.

Outliers within the unlabeled dataset were determined based on the distribution of sentence length. This distribution was then broken down into interquartile ranges, and the dataset was reduced to contain only sentences within the Q1 and Q3 ranges, which contained the middle interquartile range of data. This was done to prevent incomplete and run-on sentences from occurring within the dataset.

The generation of numeric features helps to improve the training and performance of the Random Forest model by providing structured, quantifiable representations of the text data. By breaking down the text into components such as verbs, adjectives, and nouns, the model can more effectively understand and differentiate between key characteristics of each sentence (El-Morr et al., 2022). Numeric features were generated for the training of the Random Forest model within the weak-supervision pipeline. All but 1 of the 119 numeric features were extrapolated by using the Spacy library to break down the subcomponents of the text data within each sentence. These numeric features included the number of verbs, adjectives, nouns, etc. These features were then normalized to maintain a similar scale across all features. The character length of a sentence was generated as a separate function outside of Spacy. By converting the text into numeric subcomponents, we enable the model to interpret and analyze the data effectively. Essentially, this process distills the sentences into a structured format that captures linguistic patterns, allowing the Random Forest model to operate on the underlying structures of the English language.

To process the text itself, we implemented a regex function to keep only Alphanumeric characters and a function to correct occurrences of words becoming conjoined to previous words using a Python package called Word Ninja. We set a default threshold of 6 characters within the function based on our examination of the character length distribution from all tokens in the human-annotated dataset and some trial-and-error evaluations over a select subset of sentences directly related to the issue of conjoined words.

**Machine Learning Model**

Our approach to ML development was intentionally iterative and progressive to ensure robust model accuracy and gradual complexity in design (Goodfellow et al., 2016; Xin et al., 2018). Starting with simpler NLP models, such as Bag-of-Words and n-gram models, provided essential baselines. These models allowed us to evaluate performance with low computational requirements, making it easier to identify areas for improvement before scaling up to more complex frameworks (Manning et al., 2009; Tao et al., 2021). Additionally, this stepwise progression helped us establish foundational insights, enabling better comparisons and refinements as we introduced advanced models, aligning with best practices in ML development (Chollet, 2017). We explored the use of SetFit as well as Random Forest models utilizing extracted numeric values from the text using Spacy. However, given the complexity of pulling leadership qualities from the LOR sentences, we eventually turned to Transformer-based models starting with the original BERT (Rogers et al., 2021). In training BERT, we discovered both a greater level of performance and a bottleneck pertaining to the availability of data as stated in the literature (e.g. (Niu et al., 2023)).

Our approach aimed to creatively and pragmatically enhance model performance by experimenting with BERT-based frameworks and optimizing data utilization. Initially, we generated synthetic data to increase data diversity and volume, especially to balance the minority label, 'leadership.' However, this attempt yielded limited success, as the synthetic samples did not sufficiently improve model performance (Frid-Adar et al., 2018). We then experimented with integrating a Generative Adversarial Network (GAN)-BERT framework, which combines BERT with GANs to address data scarcity issues, but this also resulted in suboptimal outcomes for our dataset (Zhu et al., 2023). In response, we turned to larger, more robust iterations of BERT, specifically using RoBERTa, which is designed to improve upon BERT's language masking and training efficiency through a more extensive pre-training process (Liu et al., 2019). RoBERTa demonstrated significant improvements over previous attempts, aligning well with the specific task. Nonetheless, we continued to explore further enhancements in pursuit of even greater performance. By iteratively refining our approach and leveraging the larger model's capabilities, we gained deeper insights into model fine-tuning and the limits of data augmentation strategies (Devlin et al., 2019).

Understanding the data bottleneck, we decided to integrate Weak Supervision techniques to create a larger pool of data from our extensive set of unlabeled data. Weak Supervision, which involves labeling data with potentially noisy annotations from multiple sources, is a widely used approach for leveraging large amounts of unlabeled data when manual labeling is costly and time-consuming (Ratner et al., 2017). Though we anticipated that a weakly supervised dataset would contain some noise, we hypothesized that the increased volume of examples could enhance the model's ability to generalize by exposing it to a broader range of data patterns (Zhou, 2018).

To implement this, we developed a custom script to generate a weakly supervised dataset. This approach allowed us to apply labeling functions and heuristics to approximate labels for unlabeled instances, maximizing the utility of our available data while balancing potential noise with the benefits of increased data diversity. Previous studies have shown that, despite some noise, weakly supervised datasets can significantly improve model performance by approximating real-world data distributions, which makes models more resilient and robust to variation (Bach et al., 2017; Zhu et al., 2023). By adopting Weak Supervision, we aimed to create a more robust dataset that would support further model fine-tuning and contribute to a better-performing final model.

During the development of the weak-supervision pipeline, confidence thresholds of 0.7 were established for both the Sentence Transformers for Few-shot Learning (SetFit) (Tunstall et al., 2022) & Robustly Optimized BERT Approach (RoBERTa) (Liu et al., 2019) models. Increasing the threshold beyond this level led to decreased performance of the models. After some trial and error, a threshold of 0.7 was determined to be effective at maintaining consistent output from the models as well as preventing the models from contributing to the pipeline on sentences where they perceived lower confidence in determining the correct label. SetFit & RoBERTa have the most extensive coverage over the unlabeled dataset by far, which led to the implementation of thresholds as a potential safeguard against undue influence over the other contributing labeling functions. The Random Forest model was initially set to have a threshold of 0.8, but due to insufficient coverage of the unlabeled dataset, the threshold was ultimately left out of the process.

The final ML model was generated using the resulting weakly supervised dataset from the previously mentioned process. RoBERTa was implemented for the final model due to its robust pre-training data having proven effective for our use case. Our dataset contains over 250k rows of weakly labeled leadership sentences. Initially, the model was trained on data subsets at intervals of 5k, 25k, 50k, & 100k. With each increase in data, the performance of the final model improved. We used the entire dataset to train the model to achieve strong performance in leadership classification within the LORs.

**LLM Model**

Our RoBERTa model analyzes the data to extract leadership-related sentences from the LORs. Building on these results, we aimed to further enrich the extracted insights. However, due to the constraints of limited annotated data and the need for deeper analysis, we integrated LLMs to augment the application's capabilities. This addition allowed us to leverage the advanced contextual understanding of LLMs to capture more nuanced details and provide a comprehensive analysis.

Since an LLM is trained on an extremely large dataset, instead of reasoning about the task at hand, it is widely known that LLMs heavily focus on extracting relevant information from the data it was trained on (Kambhampati, 2024). However, current literature on this topic suggests that there are methods through prompt engineering to get the LLM to demonstrate and apply reasoning skills (Qiao et al., 2022).

Our preliminary findings indicate that the simple approach of trusting an LLM to extract the correct phrases is not the best way to tackle this problem, as it extracts many irrelevant phrases. Recognizing the possibility of unpredictable outputs from LLMs, we implemented constraints to the generated content produced by the LLM using an external library called Guidance. Constraining the outputs of the generative language model provided the overall system with reliable consistency. This, in turn, facilitated our ability to create a pipeline from one output to another (having removed a large part of the unpredictability of the LLM outputs).  Due to the lack of additional annotated data per the subcomponents of leadership (communication, teamwork, innovation), we decided to add verification and traceability components to our pipeline. These processes were implemented using Reasoning and Acting (ReAct), a general paradigm that combines reasoning and acting with LLMs with the added capability of utilizing external tools (Yao et al., 2022). ReAct prompts LLMs to generate verbal reasoning traces and actions for a task. Essentially, it provides a way to trace the chain of thoughts or the cognitive process within the LLM, from initial reasoning to final action (Yao et al., 2022). In addition to the traceability this framework provides, it also allows for the use of external tools outside the context of the LLM model. These external tools are chosen dynamically based on the decision-making of the LLM itself.

When prompted using the ReAct framework, as seen in Figure 2, the LLM begins by generating a "Thought" related to the question, evaluating which action to take next. It then moves to the "Action" stage, where it selects and applies a predefined tool (located outside the LLM prompt). Following the use of the tool, the LLM enters the "Observation" stage, where it reports the information discovered. This process repeats iteratively until the LLM reaches a "Thought" that it has found the answer, followed by an "Action" to conclude the process and provide the final output.

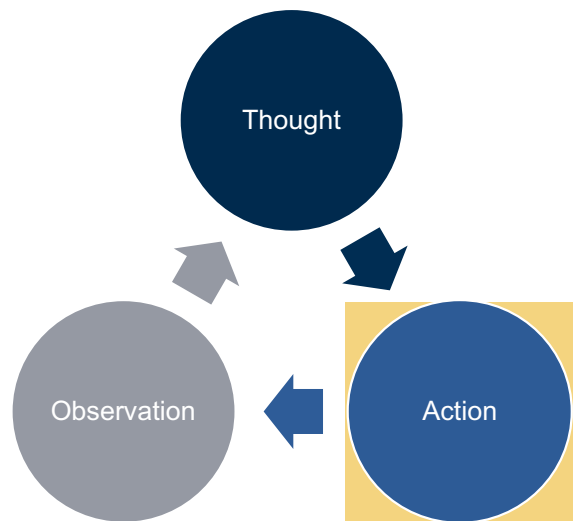

**Fig. 2** ReAct flow

To leverage ReAct, we built a separate pipeline with different prompts for each of the leadership skills we wanted to extract (teamwork, communication, and innovation). In each of these pipelines, we first used ReAct (Yao et al., 2022) practices to prompt the LLAMA2 (Touvron et al., 2023) model to generate verbal reasoning traces and actions for the task at hand. This allowed the system to perform dynamic reasoning to create and adapt plans for acting to extract teamwork, communication, and innovation phrases. Figure 3 demonstrates an example of how we utilized ReAct prompting to extract teamwork skills.

The ReAct framework provided a key advantage by enabling interaction with external tools and the environment, facilitating the retrieval and integration of additional information necessary for completing a given task. This functionality became particularly important in our work when refining and verifying the leadership phrases generated by the LLM.

**Example 1: He is an excellent communicator and a skilled collaborator when working on teams.**

| Thought 1 | I should first extract phrases which contain skills related to teamwork. |
| Action 1 | carry out thought 1 to extract phrases and separate multiple phrases using a ";" |
| Observation 1 | excellent communicator; skilled collaborator |
| Thought 2 | Are all the extracted phrases actually related to teamwork skills? I should verify each of the extracted phrases. |
| Action 2 | verify_teamwork("excellent communicator; skilled collaborator") |
| Observation 2 | excellent communicator is a teamwork phrase; skilled collaborator is a teamwork phrase |
| Thought 3 | I now know the final answer. |
| Final Answer | excellent communicator; skilled collaborator |

**Fig. 3** Sample example of ReAct prompt

To take full advantage of this capability, we incorporated an additional instance of a separate LLM model. The purpose of this separate instance was to function as a verification mechanism. Importantly, this instance was isolated from the context of the main LLM, meaning it did not have access to the ongoing prompt and responses within the original LLM session. Instead, its role was exclusively to assess and verify the phrases extracted by the primary LLM during the initial stages of the process.

The verification LLM would receive only the extracted phrases as inputs, free from any contextual biases or incomplete information from the original task. This isolation allowed for a more objective assessment, reducing the risk of errors or inconsistencies being propagated through the pipeline. By utilizing this secondary LLM model in a verification capacity, we ensured that only validated and reliable phrases were considered as the final output of the process.

## Findings

We achieved strong performance from the weakly-supervised RoBERTa model, with results indicating high accuracy and reliability. Specifically, the model attained an F1-Score of 91.6%, supported by a precision of 92.4% and a recall of 91.6%, evaluated across 524 instances in the test dataset. These metrics suggest balanced performance, demonstrating the model's effectiveness in identifying relevant instances while maintaining a low error rate.

However, an error analysis revealed that the model currently generates more false positives than false negatives. This indicates that while the model is highly sensitive in detecting relevant phrases, it tends to occasionally misclassify non-relevant instances as positive. This is likely due to overlapping features between positive and non-positive examples in the dataset. Consequently, the model over-predicted the number of leadership sentences, resulting in inter-rater reliability scores of 40.4% and 35.2% for each annotator, respectively.

Ideally, Type I errors are more acceptable in this context, as they contribute to identifying leadership qualities. However, refining the model to address its optimistic bias is an ongoing aspect of our research. Our ultimate goal is to align the model's inter-rater reliability scores with those of human-to-human Cohen's kappa metrics. The confusion matrix (Figure 4a) further illustrates the model's performance across both classes, showing a substantial number of correctly identified true positives and true negatives (240 and 244 instances, respectively). This balance highlights the model's overall effectiveness while also pointing to opportunities for fine-tuning to reduce false positives in future iterations.

Moreover, Figure 4(b) presents the summary metric for the precision-recall curve. An average precision of 0.86 confirms the high performance indicated by the confusion matrix, demonstrating the classifier's capability to effectively distinguish between positive and negative samples.

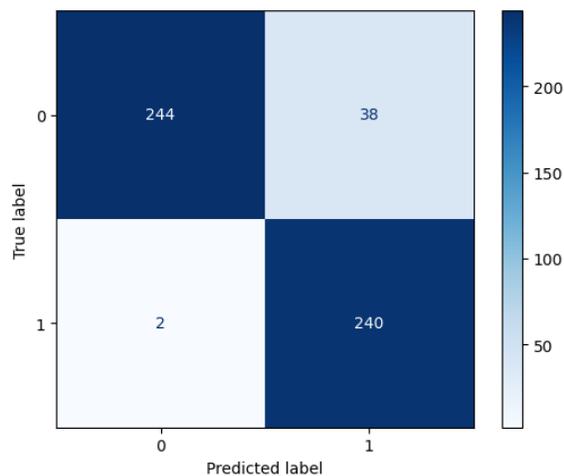
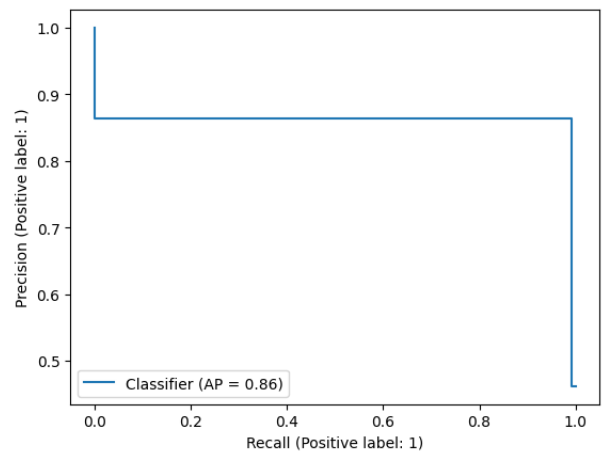

**Fig. 4 (a).** The model confusion matrix.     **Fig. 4 (b).** The precision recall curve.

Regarding LLM, one of the key components of our approach was the implementation of a verification layer using a secondary LLM. This verification LLM received only the extracted phrases as inputs, independent of any contextual information from the original task. By isolating these phrases from their broader context, the verification process mitigated potential biases and incomplete information, resulting in a more objective assessment. This strategy reduced the risk of errors or inconsistencies propagating through the pipeline, as only validated and reliable phrases were retained for the final output.

Additionally, the integration of the ReAct framework proved essential in facilitating this validation step. The framework's ability to interact with external systems allowed us to incorporate an additional LLM instance dedicated to verification, introducing a layer of independent scrutiny. This multi-step approach enhanced both the accuracy and reliability of the extracted phrases, as evidenced by the improved consistency and quality of the final outputs. The validated phrases were then used in subsequent analyses, contributing to a more robust and credible set of findings.

To effectively present applicants' leadership attributes from LORs, we developed a minimum viable product called LORI—an AI-driven web application prototype built with Streamlit in Python. As shown in Figure 5, LORI integrates multiple ML models and AI techniques, working in tandem to extract and display meaningful insights from applicants' LORs. The application accepts a PDF file containing three LORs for a given student, converts the letters into images, and applies optical character recognition (OCR) to accurately interpret and process the text. To enable seamless integration, we created additional Python scripts allowing LORI to interact with both the RoBERTa model and the LLAMA2 model (7 billion parameter version). The LOR PDF files are parsed and converted into text, which is processed by the RoBERTa model. The model's output is visualized through highlighted sections, indicating where leadership-related content is detected. These highlighted sentences are further analyzed using LLM pipelines for advanced information extraction, including phrase identification, detailed breakdown of leadership subcomponents, and an overarching summary of leadership qualities across all three LORs.

LORI demonstrates how the AI-based model performed on the tasks of detecting phrases of leadership attributes and tallying the instances of leadership-related phrases. As shown in Figure 6, the LORI provides information about the number of leadership sentences detected across multiple LORs for an individual applicant. The user can select one of the collected LORs from the dropdown menu to view results associated with the selected letter. For each selected LOR, LORI shows the full text, highlighting specific sentences that contain the leadership phrases. LORI also captures the proportion of the highlighted sentences out of the total number of sentences.

Additionally, powered by the LLM, the Summary feature offers a concise summary of the applicant's leadership attributes based on the synthesis of the information gathered across the three different LORs. We provided the LLM with leadership phases and prompted it to generate a brief overview (approximately 100 words) of each applicant's leadership qualifications. The resulting summaries are presented in this section for admission officers to reference quickly.

Furthermore,  LORI displays a bar chart that visualizes the distribution of specific attributes of leadership, including teamwork, communication, and innovation (i.e., micro-label), as illustrated in Figure 7. These results exhibit the usefulness of LLM in capturing nuanced leadership skills by drilling down into deeper details beyond the initial classifications and dramatically minimizing the data processing for phrase extraction and summarizing.

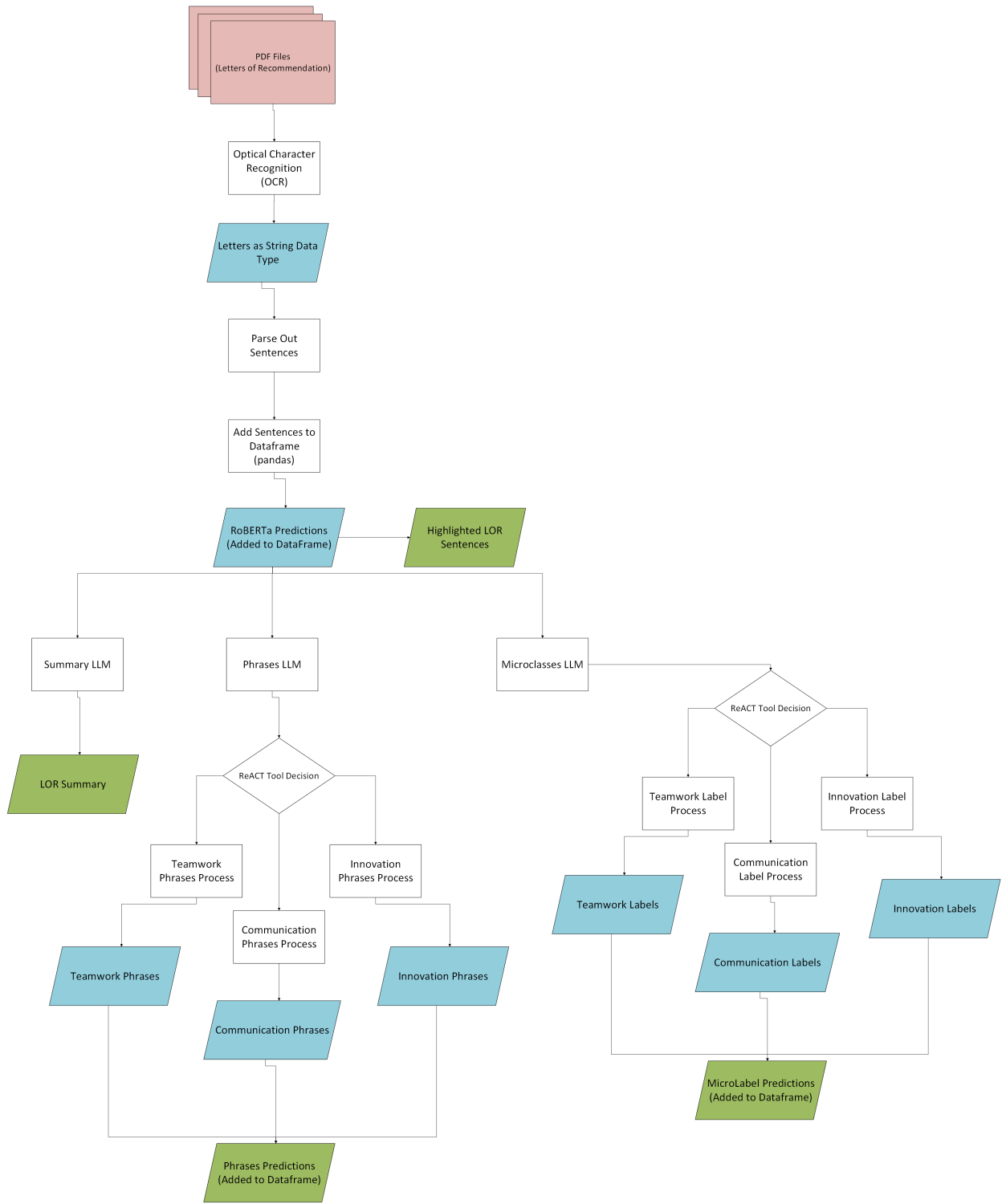

**Fig. 5** Flowchart showing LORI's process from pdf file to outputs

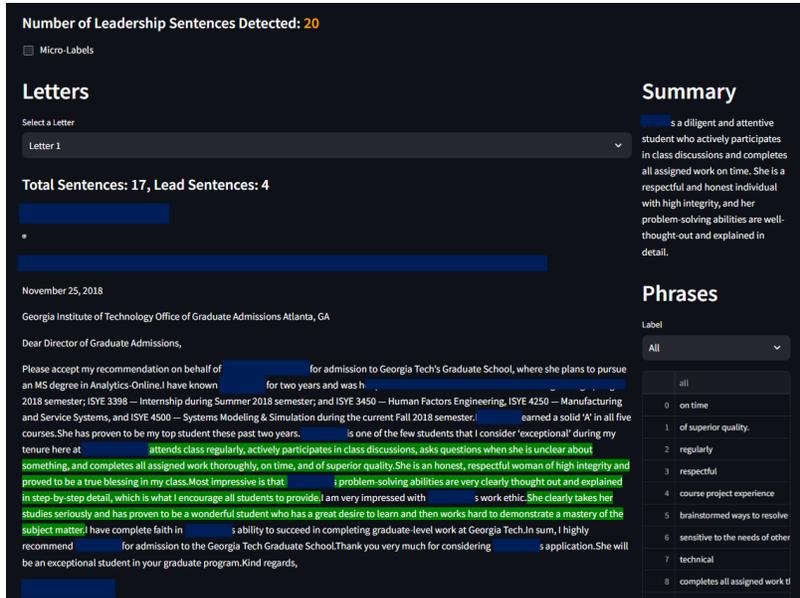

**Fig. 6** A screenshot of LORI illustrating an example of the model results.

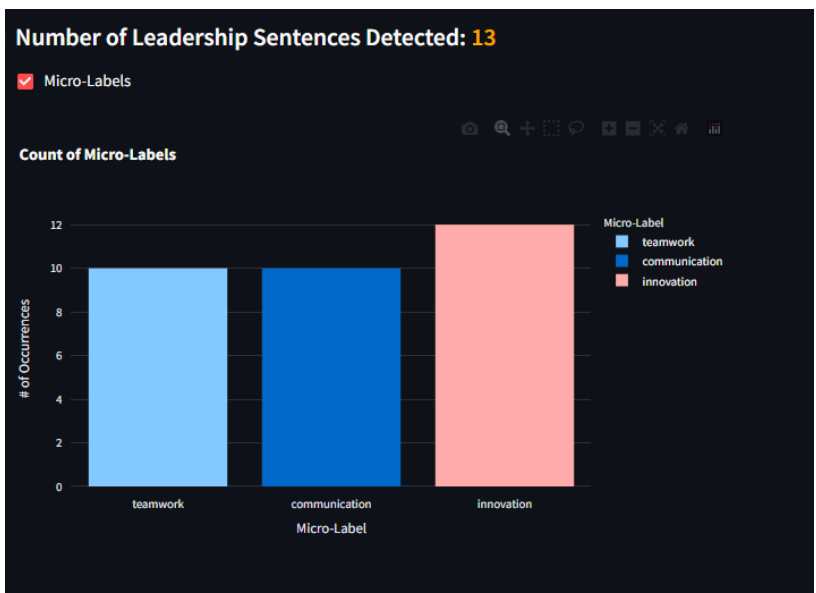

**Fig. 7** A screenshot of LORI illustrating an example of the micro-label results.

## Discussion

The RoBERTa model's overall performance on the test data was very promising and showcases the model's ability to produce a strong level of consistency in detecting leadership skills. However, we believe it is important to note that on the dataset designed to measure agreement between human annotators and the model, there was a larger pool of leadership sentences detected by the model than by the human annotators. This indicates a key point of concern: if the expert annotations are treated as proper labels for the context of the dataset, it shows that the model is biased towards positive identifications of leadership sentences. For the purposes of the LORI app, finding too many leadership skills is preferred over finding too few. However, this result likely indicates that there may be excessive noise within the weakly labeled dataset, suggesting the need for additional examination to further improve model performance. Other potential avenues for improvement include adjusting the model thresholds set for the weakly supervised dataset and possibly adding additional models to enhance the weak supervision pipeline.

Furthermore, the model's performance may be influenced by the inherent biases present in LORs (Dalal et al., 2022; Grimm et al., 2020). These biases can stem from various factors, such as the writer's perspective or the socioeconomic background of the applicant, potentially impacting how leadership attributes are described. Therefore, addressing such biases in model training and ensuring that the model generalizes well across different contexts are crucial for future development. While the RoBERTa model has shown effectiveness in this study, future research could explore integrating more diverse training data or employing hybrid models. Specifically, it may be possible to fine-tune the RoBERTa model (or a different transformer-based model) using a publicly available dataset that has a close similarity to the topic of leadership skill detection, which could benefit the model through additional data for fine-tuning.

Another promising direction involves aligning the "importance scores" of tokens within a given sentence with human perceptions of word relevancy to leadership. ML models do not weigh tokens in a sentence the way humans do (Rogers et al., 2021; Wallace et al., 2019). Consequently, developing a model that closely aligns with human perceptions would likely not only perform more effectively but also more easily extract key terms directly from the documents rather than relying on an extended process later in the pipeline (Chancellor, 2023; Truong & Koyejo, 2025). At present, the model outputs an overall summary indicating the presence of leadership traits in each letter but does not specify which terms or phrases drive the classification. In the future, improving this output to highlight the most contributive tokens could clarify the aspects of the text that signal leadership qualities. For instance, Figure 8 provides a Shapley Additive Explanation (SHAP) (Mangalathu et al., 2020; Nohara et al., 2022) output displaying feature attributions, where each token or phrase is analyzed to show its respective influence on the final prediction, such as the attribution label. By highlighting the importance levels associated with individual tokens or phrases, these explanations offer insights into which features are most impactful, thereby enhancing the interpretability of the model's predictions.

**Legend:** ■ Negative □ Neutral ■ Positive

| True Label | Predicted Label | Attribution Label | Attribution Score | Word Importance |
|---|---|---|---|---|
| 1 | LABEL_1 (1.00) | LABEL_1 | 3.88 | #s The amount of work that gets done increases when Ming is in the picture hes a much focused individual and that quality tends to rub off on his colleagues #/s |

**Fig. 8 (a)** Example of feature attribution for a true positive prediction

**Legend:** ■ Negative □ Neutral ■ Positive

| True Label | Predicted Label | Attribution Label | Attribution Score | Word Importance |
|---|---|---|---|---|
| 0 | LABEL_0 (1.00) | LABEL_0 | -1.00 | #s Her senior year project was to improve the reliability of motors employed in high speed trains #/s |

**Fig. 8 (b)** Example of feature attribution for a true negative prediction

As seen in Figure 8(a), the tokens 'amount of work that gets done increases,' 'picture,' and 'focused' are particularly influential for this specific observation from the dataset. These tokens indicate elements that the model considers significant in shaping the prediction for this instance. The phrase "amount of work that gets done increases" likely signals an emphasis on productivity or effectiveness, while "picture" may indicate the presence of the individual in a team. The term "focused" suggests an orientation toward concentration or goal-driven behavior. Together, these influential tokens suggest that the model attributes importance to themes of leadership, contribution to teamwork, and goal-driven, which cumulatively impact the final prediction outcome. Figure 8- (b) presents a case with a true negative prediction. By examining the various influential tokens, we can observe the reasons behind the negative label assignment given these attributions. In cases where the model predicts a false positive or false negative, we can observe the word attributions produced by the model to better understand how the model arrived at that conclusion, which in turn informs us, researchers, on additional considerations when developing and improving the model. Overall, these attributions will pave the way to understanding the nuanced factors that the model interprets as key drivers for the final prediction.

Moving forward, we plan to enhance the RoBERTa model's performance by employing Bayesian Optimization for hyperparameter tuning. Bayesian Optimization is an effective method for hyperparameter search, utilizing a probabilistic surrogate model to explore the parameter space efficiently with fewer evaluations (Frazier, 2018; Snoek et al., 2012). By iteratively converging on an optimal set of hyperparameters, this approach could significantly improve the model's predictive accuracy and generalizability (Wu et al., 2017).

Additionally, the LLM component of our tool is undergoing further investigation to assess its effectiveness in tasks such as summarization, phrase extraction, and micro-label categorization. LLMs, including GPT and BERT-based models, have demonstrated strong capabilities in generating high-quality text summaries and extracting meaningful phrases due to their advanced contextual understanding (Brown, 2020; Raffel et al., 2020). However, their performance is sensitive to factors such as model size, architecture, and tuning parameters (Devlin et al., 2019). To address this, we aim to refine evaluation metrics that encompass both qualitative and quantitative dimensions. This will allow us to comprehensively assess the model's capabilities. These ongoing efforts will enhance the robustness and scalability of our tool, ensuring its effectiveness in real-world applications.

The LORI dashboard is a pivotal component of the AI-driven system designed to assess leadership qualities in applicants through their LORs. This dashboard offers a user-friendly interface that provides evaluators with clear, actionable insights into an applicant's leadership attributes, thereby streamlining the admissions process. Notably, the dashboard employs intuitive visualizations to highlight identified leadership qualities such as teamwork, communication, and innovation. This visual strategy allows admissions committees to quickly understand an applicant's strengths and areas for development. Users can also explore specific sections of the LORs to gain deeper context into how leadership traits are put forward. As a result, this interactivity helps admissions committees make well-informed decisions based on comprehensive data. The dashboard enables easy comparison between applicants by aggregating leadership sub-scores in teamwork, communication, and innovation and presenting them side by side. This helps identify standout candidates and supports a fair evaluation process.

To further enhance the dashboard, additional text analytics features like word count, readability scores, and word clouds could be integrated. For example, readability scores can indicate the complexity and accessibility of the text, while word clouds provide a visual representation of the most frequent terms, offering a quick overview of key themes (DePaolo & Wilkinson, 2014; Kalmukov, 2021). Moreover, by setting a baseline for average metrics across the student population, the tool could enable comparative analysis of leadership skills in individual students. Research suggests that comparative analytics can provide meaningful insights, facilitating personalized feedback and helping educators identify unique student strengths and areas for improvement (Bernacki et al., 2021). Such enhancements would contribute to a more thorough evaluation of student competencies and support targeted educational interventions.

It is essential to recognize that recommendation letters are generally selected and written by referees specifically to reflect a positive assessment of an individual's skills, achievements, and personality. This selection process is inherently biased, as individuals tend to choose referees who are likely to portray them favorably (Gillis, 2021; Morgan et al., 2013). As a result, this leads to selection and representation biases, with referees more likely to highlight strengths while minimizing weaknesses, creating a skewed sample that is not representative of all possible opinions on an individual's character and abilities (Chapman et al., 2022).

Gender differences further complicate these biases in LORs. Studies have shown that recommendation letters for men often emphasize accomplishments, leadership, and intellectual ability, whereas those for women tend to focus more on personal attributes like kindness or diligence, using more subjective, relationship-oriented language (Dutt et al., 2016; Madera et al., 2019; Madera et al., 2009; Schmader et al., 2007). This language bias can disadvantage women by aligning with stereotypical gender roles rather than objective measures of qualifications. Furthermore, omitted information bias (Chapman et al., 2022) presents another concern. Given that LORs are typically brief, critical details may be omitted, either intentionally or unintentionally. This is problematic, as ML models only have access to the provided content, lacking insight into any significant missing information. For instance, a referee may omit notable achievements of a female applicant due to implicit gender biases, leading the model to undervalue her qualifications compared to male counterparts.

Therefore, if these biases are not addressed, they could result in systematic errors in candidate evaluation, favoring traits often highlighted in LORs for male applicants. To address these issues, adversarial training —a technique that exposes the model to specially designed examples to help it recognize and correct biased patterns—can help the model distinguish between biased and unbiased representations. Moreover, explainable artificial intelligence (XAI) techniques can be used to assist in identifying influential features that drive predictions, which will contribute to promoting a fairer assessment process (Adadi & Berrada, 2018; Gunning et al., 2019). By implementing these strategies to mitigate the effects of gender biases, the evaluation of LORs can become more equitable.

## Conclusion

The increasing emphasis on leadership skills in graduate education underscores the need for innovative solutions. Developing and validating LORI to detect these skills benefits both admission committees and applicants by automating the review process, significantly reducing the time and effort required to evaluate application documents. This automation leads to more precise and efficient admissions decisions. As higher education shifts toward holistic reviews that prioritize a broader set of candidate qualities, LORI emerges as an essential tool to promote equity, efficiency, and depth in the admissions process—ultimately shaping a more competent, adaptable, and diverse future workforce.

Beyond admissions, LORI's versatility extends to instructional settings, where it can analyze other educational data sources such as essays, peer evaluations, and project reflections. By integrating LORI into formative assessment, peer feedback, project-based learning, and virtual classrooms, institutions can enhance leadership development in a scalable, data-driven manner. As a formative assessment tool, LORI provides students with automated, targeted feedback on leadership development through essays, discussion posts, and reflections, allowing for early identification of strengths and areas for improvement while fostering personalized learning paths. In peer review processes, LORI facilitates structured feedback by analyzing evaluations, identifying leadership traits in student submissions, and encouraging self-reflection. Within project-based learning, it assesses leadership indicators in team reports and reflections, enabling faculty to track leadership growth and identify emerging student leaders.

Looking ahead, LORI's applications could expand further, including piloting in leadership-focused graduate courses, integrating into corporate training programs, and adapting to discipline-specific leadership needs in STEM, business, humanities, and healthcare. By embedding LORI into diverse instructional contexts, institutions can foster leadership development in a more systematic and competency-driven manner, equipping students with the essential skills needed to thrive in dynamic professional environments.

## Declarations

**Availability of data and materials**

The datasets generated and/or analyzed during this study are not publicly available to protect applicant confidentiality, but anonymized data can be obtained from the corresponding author upon reasonable request.

**Funding**

The authors declare that no funds, grants, or other support were received during the preparation of this manuscript.

**Acknowledgments**

Not applicable

**Clinical trial number**

Not applicable

**Ethics declaration**

This study was reviewed and approved by the Institutional Review Board at the Georgia Institute of Technology.

**Consent to participate**

Not applicable.

**Consent for publication**

Not applicable.